\documentclass[letterpaper, 10 pt, conference]{ieeeconf}  

\IEEEoverridecommandlockouts                              

\usepackage{graphicx}
  \graphicspath{{../figures/}}
  \DeclareGraphicsExtensions{.pdf,.jpeg,.png}
\usepackage{epsfig} 
\usepackage{mathptmx} 
\usepackage{times} 
\usepackage{amsmath} 
\usepackage{amssymb}  
\usepackage{flushend}
\usepackage{dblfloatfix}    
\usepackage[dvipsnames]{xcolor}
\usepackage{caption}
\usepackage{subcaption}
\usepackage{booktabs}

\usepackage{siunitx}
\title{\LARGE \bf
Comparison of Constrained and Unconstrained Human Grasp Forces Using Fingernail Imaging and Visual Servoing}

\author{Navid Fallahinia$^{1}$ and Stephen A. Mascaro$^{1}$
\thanks{*This work was partially supported by National Science Foundation (NSF) Grant NRI-1208626}
\thanks{$^{1}$Authors are with the Robotics Center, Mechanical Engineering Department, University of Utah, Salt Lake City, Utah, USA,
        {\tt\small n.fallahinia@utah.edu}}%
}

\begin{document}

\maketitle
\thispagestyle{empty}
\pagestyle{empty}

\begin{abstract}

Fingernail imaging has been proven to be effective in prior works \cite{grieve20153,grieve2016optimizing} for estimating the 3D fingertip forces with a maximum RMS estimation error of 7\%. In the current research, fingernail imaging is used to perform unconstrained grasp force measurement on multiple fingers to study human grasping. Moreover, two robotic arms with mounted cameras and a visual tracking system have been devised to keep the human fingers in the camera frame during the experiments. Experimental tests have been conducted for six human subjects under both constrained and unconstrained grasping conditions, and the results indicate a significant difference in force collaboration among the fingers between the two grasping conditions. Another interesting result according to the experiments is that in comparison to constrained grasping, unconstrained grasp forces are more evenly distributed over the fingers and there is less force variation (more steadiness) in each finger force. These results validate the importance of measuring grasp forces in an unconstrained manner in order to study how humans naturally grasp objects.

\end{abstract}

\section{INTRODUCTION}

The complexity of a human hand and the fact that it can reach so many degrees of freedom make it difficult to study its motor behavior during a grasp or manipulation task. There is a significant gap in our understanding of how the human brain can learn and execute these motor behaviors. In order to address this gap and understand the sensorimotor control mechanisms that are employed by the central nervous system, the force collaboration strategies among the fingers must be studied in a totally natural manner. Improvement in our knowledge of human grasp strategies may be used almost immediately in different areas, including robotics, haptics, human-robot interactions, rehabilitation, and bioengineering \cite{gailey2017grasp,polygerinos2019soft,santello2016hand,godfrey2018softhand}. Moreover, The increased understanding of precision grasping will result in more accurate mathematical modeling of dexterous grasping \cite{mojtahedi2015extraction}.

The main concern in human grasp force study is to measure the applied forces at the fingerpad on multiple fingers under natural conditions during experiments i.e., without restricting the human's haptic senses or constraining how the human grasps an object. Classical methods of studying human grasping require fingers to be placed on a set of pre-specified locations on an instrumented object where force sensors are mounted \cite{soechting2008sensorimotor,fu2010anticipatory}. In these methods, finger contact positions are defined by the locations of the force sensors along the sides of the object to be grasped. Previous studies have shown that unlike the unconstrained posture, constraining the finger placement can create a meaningful difference in force distribution over the fingers \cite{lukos2013grasping}. However these studies only involved the index finger and thumb, since their instrumented objects cannot allow for independent measurement of multiple unconstrained finger forces. Considering  the fact that humans do finger placement so efficiently, it is necessary to measure grasp forces on multiple fingers in an unconstrained manner. There have been several methods developed for unconstrained grasp force measurement \cite{tanaka2008novel,naceri2014coordination,battaglia2016thimblesense}. However, these techniques are either not capable of measuring shear forces, or may interfere with hand movements and haptic sense due their complex de,sign, and cannot be used for the purpose of studying grasp force. Thus, the goal of this paper is to conduct multifingered grasping experiments under both constrained and unconstrained conditions to study the effect of these conditions on the observed force synergies. For this purpose, we have used the fingernail imaging method, which has been developed by the authors, as a tool for measuring the forces.

\subsection{Fingernail Imaging}

The fingernail imaging technique, which has been developed in the previous works, is an effective method of 3D fingertip  force estimation without imposing any constraint on finger placement \cite{grieve20153}. For this method, forces are estimated by analyzing the coloration patterns on fingernail due to the blood flow change in tissues beneath the nail when the fingertip comes in contact with objects. This method is capable of  estimating fingertip forces along all three directions simultaneously with a maximum estimation RMS error of $0.55 \pm 0.02$ N, which was 7\% of the full range of forces measured. \cite{grieve2013force,fallahinia2018grasp,fallahinia2019grasp}
\renewcommand\thefigure{\arabic{figure}}
\setcounter{figure}{1}
 \begin{figure*}[!b]
      \centering
      \includegraphics[trim={0 0 0 -1cm}, width=0.65\textwidth]{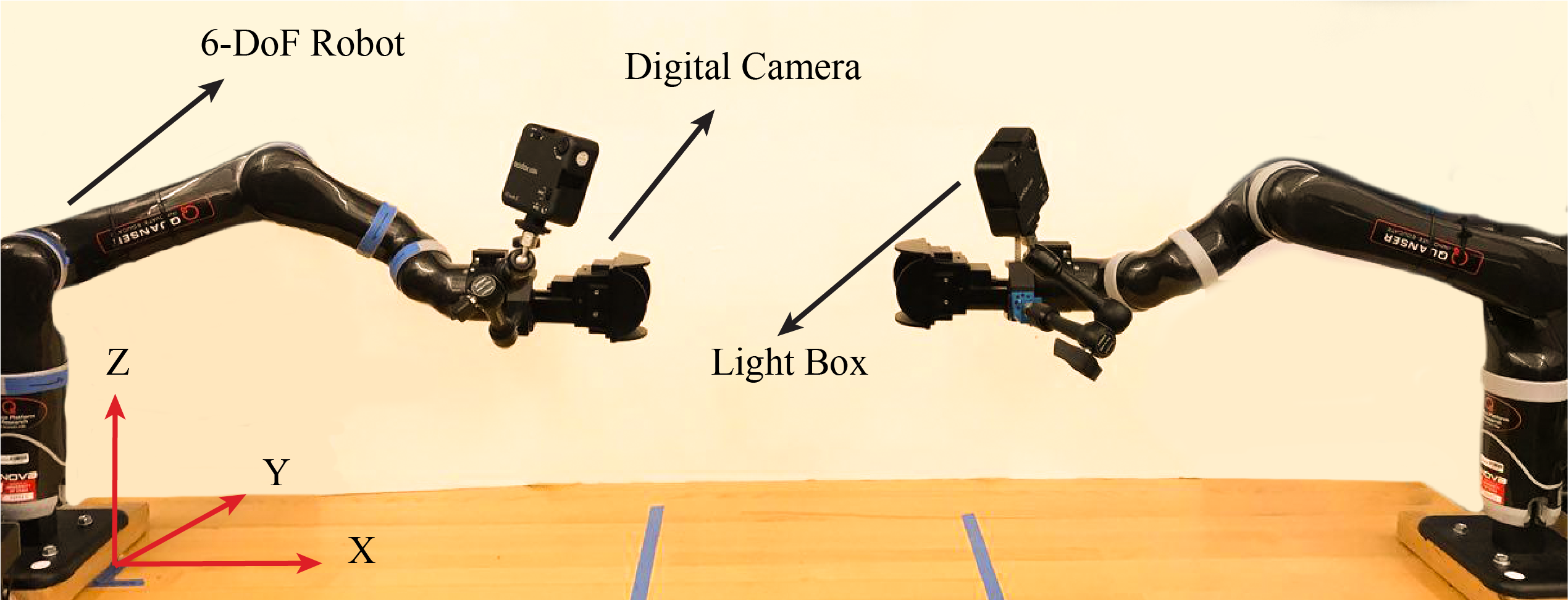}
  \caption{The Experimental Setup for grasping experiments. The setup includes two 6-DOF MICO2 Kinova robots, PointGrey Flea3 digital cameras and the light boxes attached to each robot. The global coordinate system has also been pictured. The subject will be asked to hold an object in a four fingered precision grip between two cameras. Images will be recorded using the cameras. The robots will hold each camera on each side of the hand.
}
  \label{tworobotssetup}
\end{figure*}

The capability of the fingernail imaging method is, however, limited by the ability to maintain a proper view of the human fingernails with one or more cameras as the fingernails translate and rotate within a workspace during a dynamic grasping task. To overcome this problem, cameras can be mounted on the end-effectors of robotic arms, and a visual servoing tracking system can be employed to ensure that the fingers are in the camera view all times. As a result, grasp force estimation using fingernail imaging is a two-step process (Fig.~\ref{twostageprocess}): 1) building a force estimation model based on a collected set of calibration data including fingernail images and applied forces 2) applying the force estimation model on the new images captured during the grasping experiments. These two steps will be explained in the following sections.  
\renewcommand\thefigure{\arabic{figure}}
\setcounter{figure}{0}
\begin{figure}[!t]
  \centering
  \includegraphics[trim={0 0cm 0 -1cm}, width=0.4\textwidth]{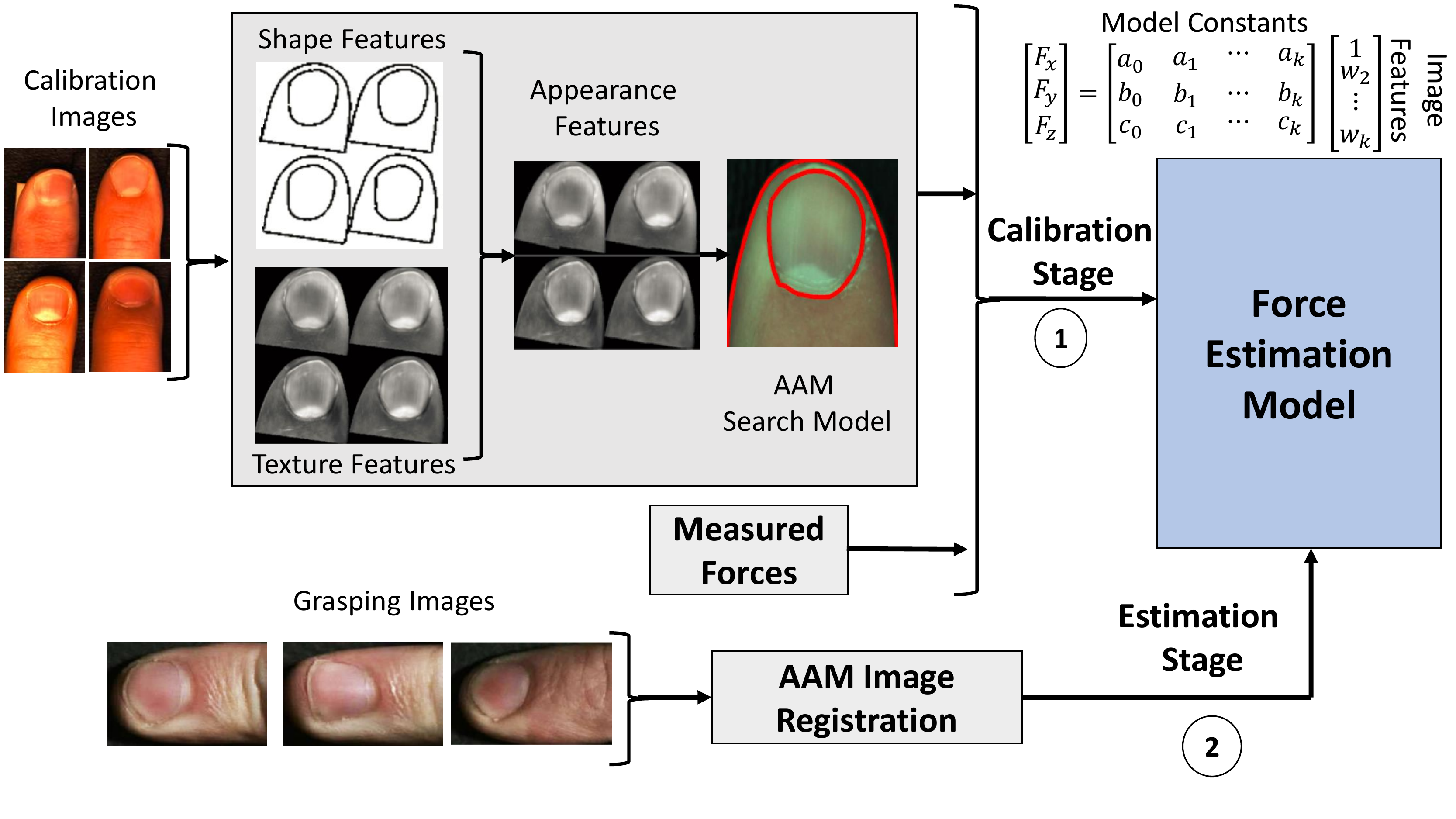}
  \caption{The two steps of grasp force estimation using fingernail imaging based on AAM registration method. 1) building the calibration model, 2) applying the force estimation model on the grasping images.}
  \label{twostageprocess}
    \vspace{-1.5em}
\end{figure}

\section{Experimental Procedure}
In order to study the grasp forces under both constrained and unconstrained grasping conditions, data from 6 human subjects (3 male and 3 female) with different fingernail size and texture was recorded.

\subsection{Training Data Collection}

The training data was collected using an automated calibration platform that was developed in the previous work \cite{grieve20153}. The data collection experimental setup includes a Magnetic Device (MLDH) that exerts desired force levels according to a hybrid position-force controller with feedback linearization. The calibration desired force space is a Cartesian grid that includes normal force between 0N to 18N and shear forces in the range -3N to 3N and -6N to 6N along x and y axis, respectively. Forces were recorded using a 6-axis force sensor, while fingernail images were captured by an RGB camera. To eliminate the ambient light and to provide uniform lighting during the data collection, an LED lightbox was installed above the human hand to remove any glare on the fingernail. Each of the index, middle, ring and thumb fingers were calibrated individually per each human subject and about 6725 to 7050 images and corresponding forces were recorded for each individual finger. 

\subsection{Grasping Experiments}

For the grasping experiments, it is desired to keep the fingernails centered in the field of view of the camera and to keep the camera at a preferential distance to the fingernails (to maximize the size of the nails in the image). Using a stationary camera, human subjects cannot be relied upon to keep their fingernails in the camera view. This restriction would limit further research. Therefore, we have employed an autonomous visual tracking system to allow a camera mounted on a robot to track the fingernails during a grasping task. Two 6-DOF robotic arms were chosen for this task. Two PointGrey Flea3 digital cameras mounted on the last link of each robot were also selected for the purpose image capturing during the grasping experiments one with a view of the thumb and the other, the fingers. Two LED light boxes were attached to each robot end-effector to provide the same lighting condition as the one used in calibration experiments. Fig.~\ref{tworobotssetup} shows the experimental setup for grasping experiments.

In order to perform grasping tests with both constrained and unconstrained conditions, two grasping objects have been created. The first object is an instrumented T-shaped apparatus with precise grasping locations, where four force sensors are used for measuring the contact forces and validating the estimated forces found by fingernail imaging (Fig.~\ref{graspingobject}(a)). Previous studies have shown that little finger forces are small enough during grasping, that they can be neglected \cite{fu2010anticipatory}. Hence, only four-fingered grasping has been studied in this research. The second grasping object, which has been designed for unconstrained grasping tests, is the same T-shaped object but, with no pre-defined grasping points, such that the fingers may contact anywhere along the pads on both sides as shown in Fig.~\ref{graspingobject}(b). Moreover, a sticky tape covered the finger placement locations on both objects to increase friction and prevent the fingers from sliding.
\renewcommand\thefigure{\arabic{figure}}
\setcounter{figure}{2}
\begin{figure}[!t]
      \centering
      \includegraphics[trim={0 0 1cm 1cm}, width=0.35\textwidth]{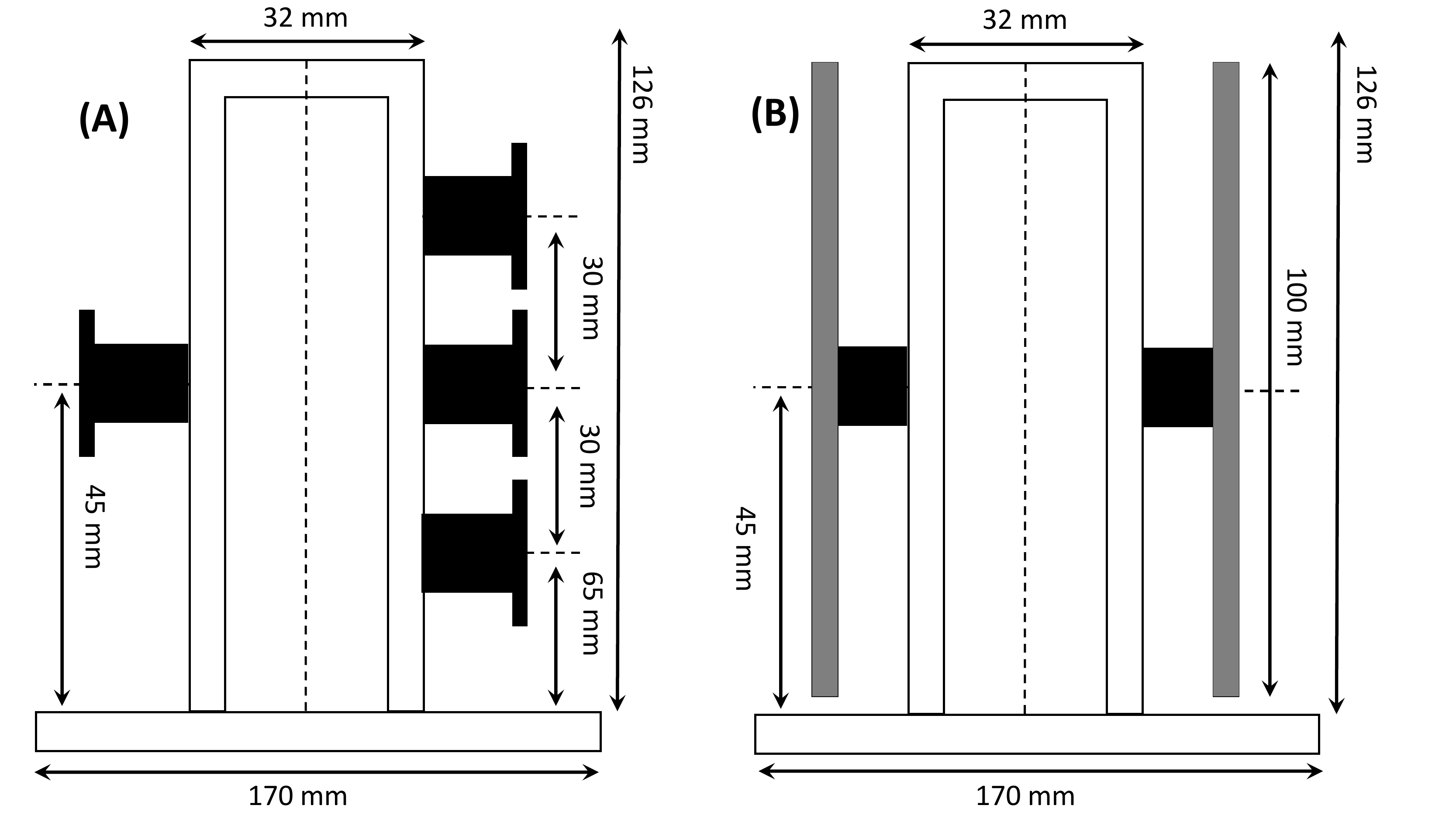}
      \caption{Grasping objects. Both objects have the same the size and equal weights (1.14 Kg). (a) an instrumented object with 4 force sensors mounted to be used in constrained grasping experiment. (b) an object with no pre-specific grasping points used in unconstrained grasping experiment.}
      \vspace{-1em}
      \label{graspingobject}
\end{figure} 

\begin{figure}[!b]
      \centering
      \includegraphics[trim={0.7cm 6cm 13cm 0cm}, width=0.4\textwidth]{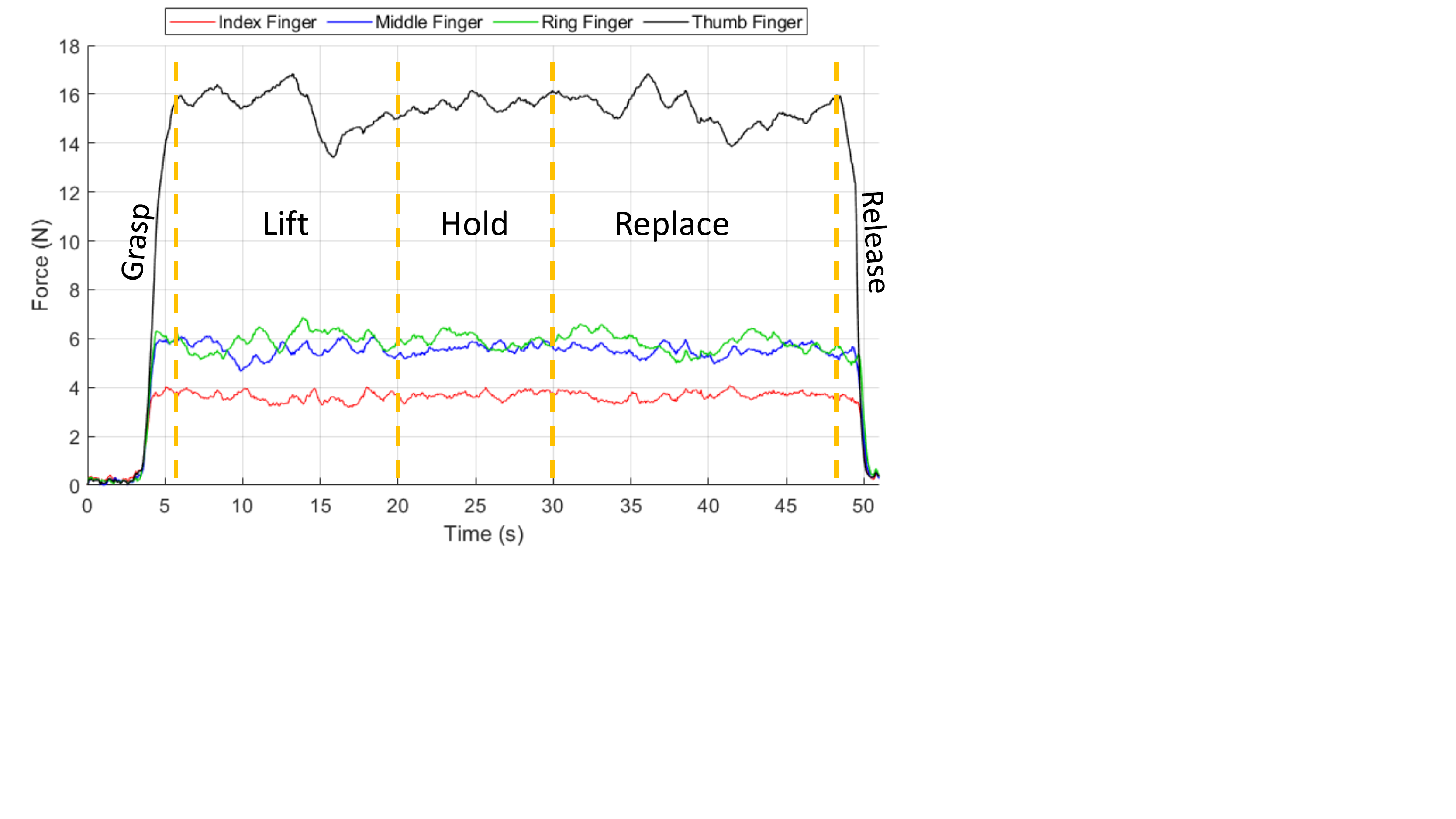}
      \caption{The normal finger forces for the entire unconstrained grasping experiment for one human subject. All four grasping phases including grasp, lift, hold, and replace are shown.}
      \label{forceovertime}
     \vspace{-1em}
\end{figure}

Two sets of grasping tests were carried out to investigate the effect of constraining the grasp, including one with the constrained object and one with the unconstrained object. In both tests, the experimental procedure for each test subject includes reaching for the object, grasping it, lifting it up to 25 cm from its initial position, holding it for 10 sec and replacing the object in its initial position. Subjects could see the instructions for each phase of the experiment as well as an online video of their fingers on a screen in front of them. First, the test subjects were instructed to stand in front of the table with their shoulder aligned with the object, and to have the corresponding hand 20-30 cm from the object. After 5 seconds, they were instructed to put their fingers on the object. After another 3 seconds, which is required to make sure that the fingernails are inside of the camera frame, the subjects were asked to lift the object. The total time needed for this experiment could vary between 45 to 55 secs per each subject and each test was repeated two times per each human subject. To give a better understanding of the four grasping phases, Fig.~\ref{forceovertime} shows the normal force per each finger over the entire grasping experiment time for one of the human subjects.

\section{Method and Data Analysis}

After data collection for each test subject from both calibration and grasping experiments, the grasp forces are estimated from the fingernail images. In this section, the steps that are needed for force estimation using the fingernail imaging method will be discussed. Moreover, the visual servoing system used for finger tracking will be explained.

\begin{figure}[!b]
      \centering
      \includegraphics[trim={0cm 9cm 0cm -1cm},width=0.45\textwidth]{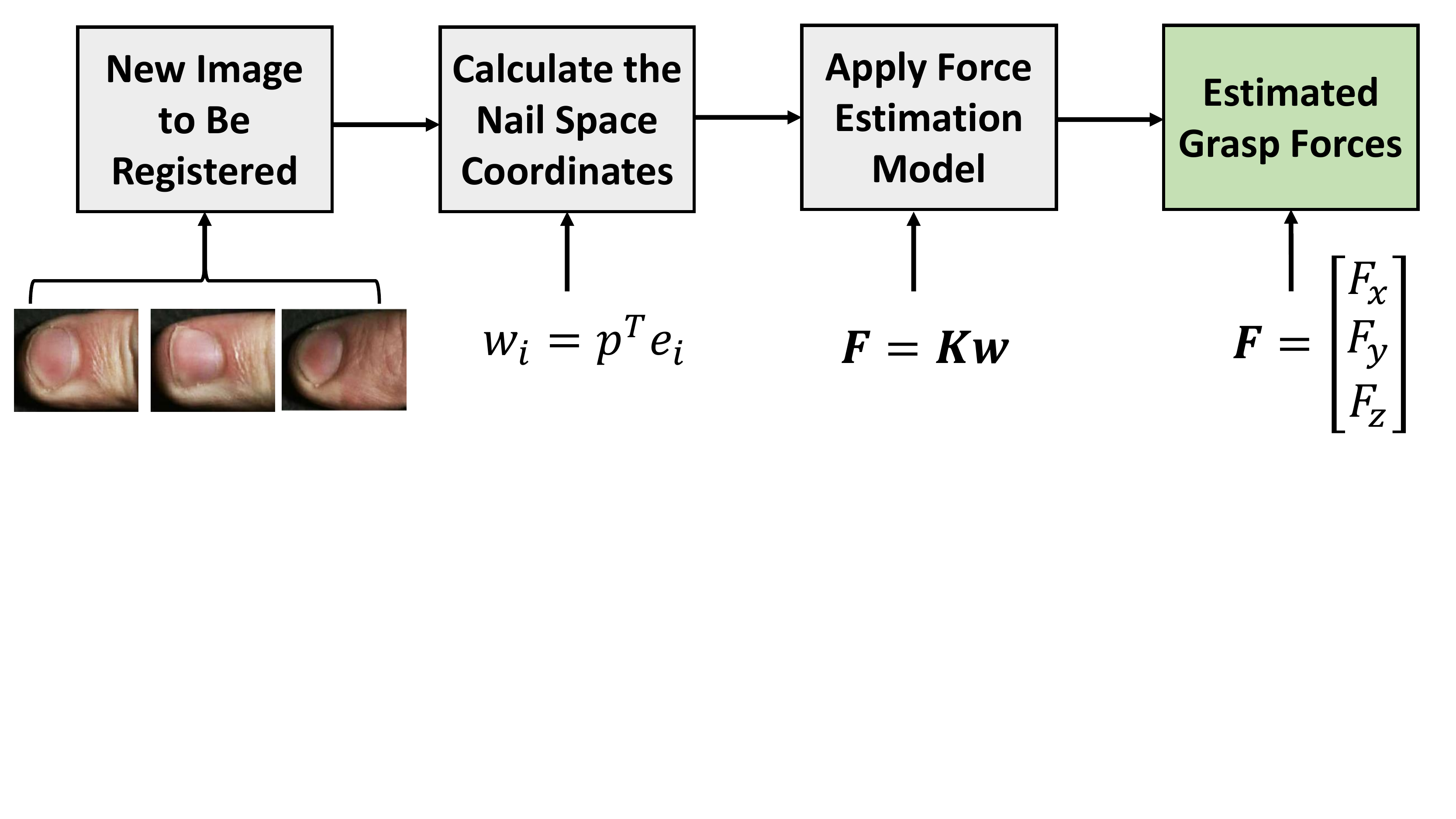}
      \caption{Force Prediction using the EigenNail Magnitude Model. The Nail Space coordinates are calculated by projecting the image pixel intensities onto each EigenNail to find the Nail Space coordinates ($w_i$). Then the prediction equation is applied, using the model constants calculated during calibration.}
      \label{EigenNail_Model_Final}
      \vspace{-1em}
\end{figure}

\subsection{Force Estimation Model}

Before building any force estimation model, all the fingernail images must be registered and warped to a template to compensate for any scaling, rotation, and translations on the images. The same registration method that was developed in previous works \cite{grieve2013fingernail} is used in this paper which iteratively uses Active Appearance Models (AAM) to register all of an individual's data. AAM is the combination of shape and texture models. Thus, each of those models should be determined separately (using Principal Component Analysis (PCA)) and then combined together. Once the shape and texture parameters vectors have been found, the AAM search model can be formed using the following equation:

\begin{equation}
    \textbf{b} =  \Phi_{c}\textbf{c}
  \label{AAM_model}
\end{equation}

\noindent Where $\textbf{b}$ is the appearance model which is the combination of shape and texture models. $\Phi_{c}$ and $\textbf{c}$ are defined in \cite{grieve20153}. For each new image chosen to be registered, the finger mean shape is found and is placed in an estimated location on the image. Then, the AAM Search Model is applied to find the final location of the finger shape. Finally, the image is warped to the template using a piecewise linear transformation. 

Once a new image is warped, the EigenNail Magnitude Model \cite{grieve2013force} is used to create a force estimation model, which forms a relating mapping between the images coordinates to the force in Nail Space. As the first step, PCA is performed on the calibration images to find the eigenvectors of the data set. The first $k$ eigenvectors, which are responsible for 99\% of the data variation, will be kept. Then, to find the Nail Space coordinates ($\textbf w = [w_1 w_2 ... w_k] $), each image is projected onto their  eigenvectors. Finally, a least-squares regression is performed to relate the Nail coordinates to the 3D forces. Fig.~\ref{EigenNail_Model_Final} shows the overall process of force estimation, where images Nail Space coordinates are used as inputs for force estimation model according to the following equation:

\begin{equation}
    \mathbf{f} =\begin{bmatrix}
  a_1 & a_2 & \cdots & a_n \\ 
  b_1 & b_2 & \cdots & b_n \\ 
  c_1 & c_2 & \cdots & c_n 
  \end{bmatrix} \mathbf{w} + \begin{bmatrix}
  a_0\\ 
  b_0\\ 
  c_0
  \end{bmatrix}
  \label{AAM_model}
\end{equation}

\noindent Where $\textbf{f}$ is the vector of estimated force, and $a_i$, $b_i$ and $c_i$ are the coefficients found by multivariable regression.

\begin{figure}[t!]
      \centering
      \includegraphics[trim={0cm 0cm 0cm -1cm}, width=0.35\textwidth]{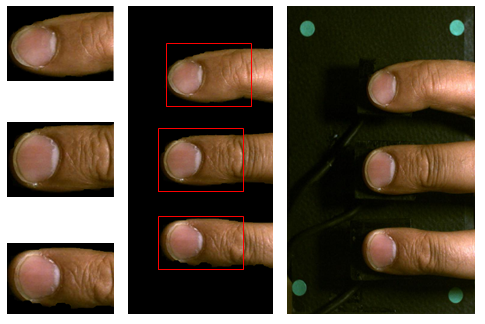}
      \caption{Finger segmentation using thresholding. All three identified and located fingers on a frame of the captured image during a grasping test. }
      \label{fingerREcog}
      \vspace{-1.5em}
\end{figure}

Since each finger has a different force estimation model, fingers must be first detected on each of the captured images from grasping tests (index, middle and ring fingers are in one image) to be used by the estimation model. For this purpose, a simple thresholding color segmentation in HSV color space is used, where fingers are detected by considering a threshold for the H values of the skin color. Finally, the images must be appropriately labeled by order of the fingers. Since the order is always index, middle, and ring (from top to the bottom of the image), it is only necessary to check the order of the detected fingers on the image and label them accordingly. After detecting and labeling each individual finger, a region with the size of $300 \times 600$ pixels around each detected fingernail is cropped and will be used as an input to the force estimation model as shown in Fig.~\ref{fingerREcog}. 

\subsection{Visual Servoing}

During the grasping experiments, a visual tracking system on each robot controls their motion using feedback information extracted from an image and is based on a mapping between the object features in the image plane and the object's motion (image-based visual servoing) to ensure that the two robotic arms can follow the fingers all times. Using the eye-in-hand setup allows the same camera that is being used to track the fingers to be simultaneously utilized to capture images of the fingernails for force estimation.

The control scheme used in this paper is based on the principles described in \cite{chaumette2006visual}. The controller is designed to use the image-space approach, which ensures that the robot will take the shortest path in image space. The image features that are used for tracking are the centers of a set of four dots that are placed on four corners of the grasping objects. The visual servoing runs at 1KHz, while the images are captured with a frequency of 20Hz with the resolution of $680\times1024$ pixels. The two robots are controlled independently, while the image capturing of the two cameras is synchronized using a TCP/IP connection between them. 

\section{Results and Discussion}

\begin{figure}[t!]
      \centering
      \includegraphics[trim={0cm 0cm 0cm -1cm}, width=0.35\textwidth]{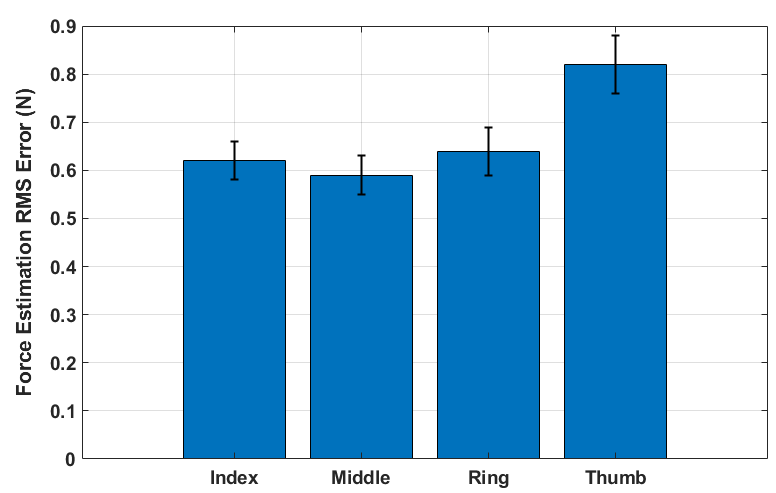}
      \caption{Finger segmentation using thresholding. All three identified and located fingers on a frame of the captured image during a grasping test. }
      \label{rmserror}
      \vspace{-1.5em}
\end{figure}

As the first result, Fig.~\ref{rmserror} shows the RMS validation error between the measured (using the force sensors) and the estimated forces in both normal and shear directions during the constrained grasping. The RMS estimation errors for each of the fingers are $0.82 \pm 0.06$ N (the thumb), $0.62 \pm 0.04$ N (index finger), $0.59 \pm 0.03$ N (middle finger), and $0.64 \pm 0.04$ N (ring finger). A Mann-Whitney \textit{u} test indicates that there is no significant difference in the errors between the estimated and measured forces among each of the four fingers ($p = 0.22$, $\alpha = 0.05$).

\begin{figure}[!b]
  \vspace{-1em}
  \centering
  \includegraphics[trim={3cm 8cm 3cm 8cm}, width=0.35\textwidth]{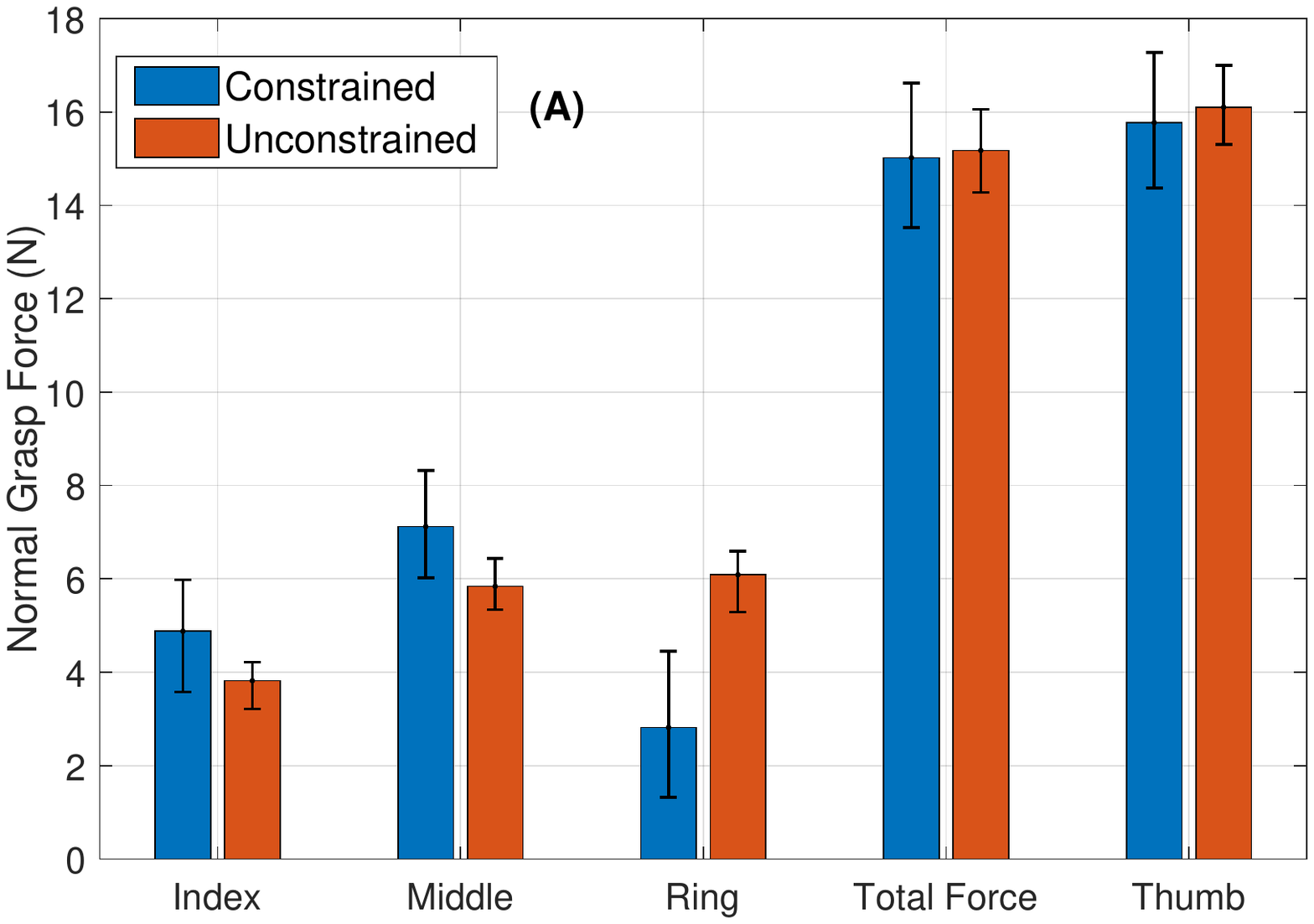}
  \vspace{-1.4em}
\end{figure}

\begin{figure}[!b]
  \centering
  \includegraphics[trim={2.5cm 8cm 2.5cm 7.85cm}, width=0.35\textwidth]{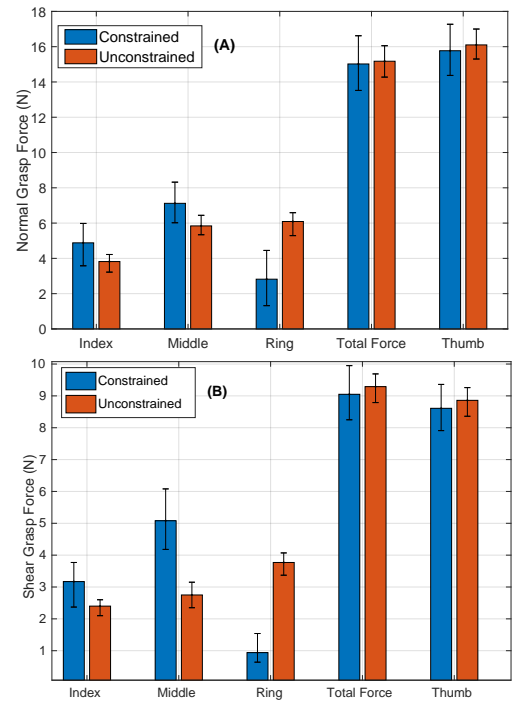}
  \caption{The average estimated forces for each finger during a 10-seconds hold phase across all the subjects (All forces are shown in absolute values). (a) The average estimated constrained and unconstrained normal forces, (b) The average estimated constrained and unconstrained shear forces.}
  \label{forcenormal}
\end{figure}

Fig.~\ref{forcenormal}(a) and (b) show the average normal and shear estimated forces for each finger during the 10 seconds hold phase. Since the forces are static during the hold phase, The total force (the overall force applied by the three fingers) must be equal to the thumb force. According to Fig.~\ref{forcenormal}(a) and (b), there is a maximum $5.7\%$ difference between the total force and the thumb force along both normal and lateral directions which validates the force estimation accuracy. Another interesting result is the force magnitude change in each finger by switching from constrained to unconstrained condition. It can be seen from the figures that the force distribution among the three fingers changes based on the grasping condition and positions of the fingers. 

\begin{figure}[!t]
  \centering
  \includegraphics[trim={2.5cm 8cm 2.5cm 6cm}, width=0.35\textwidth]{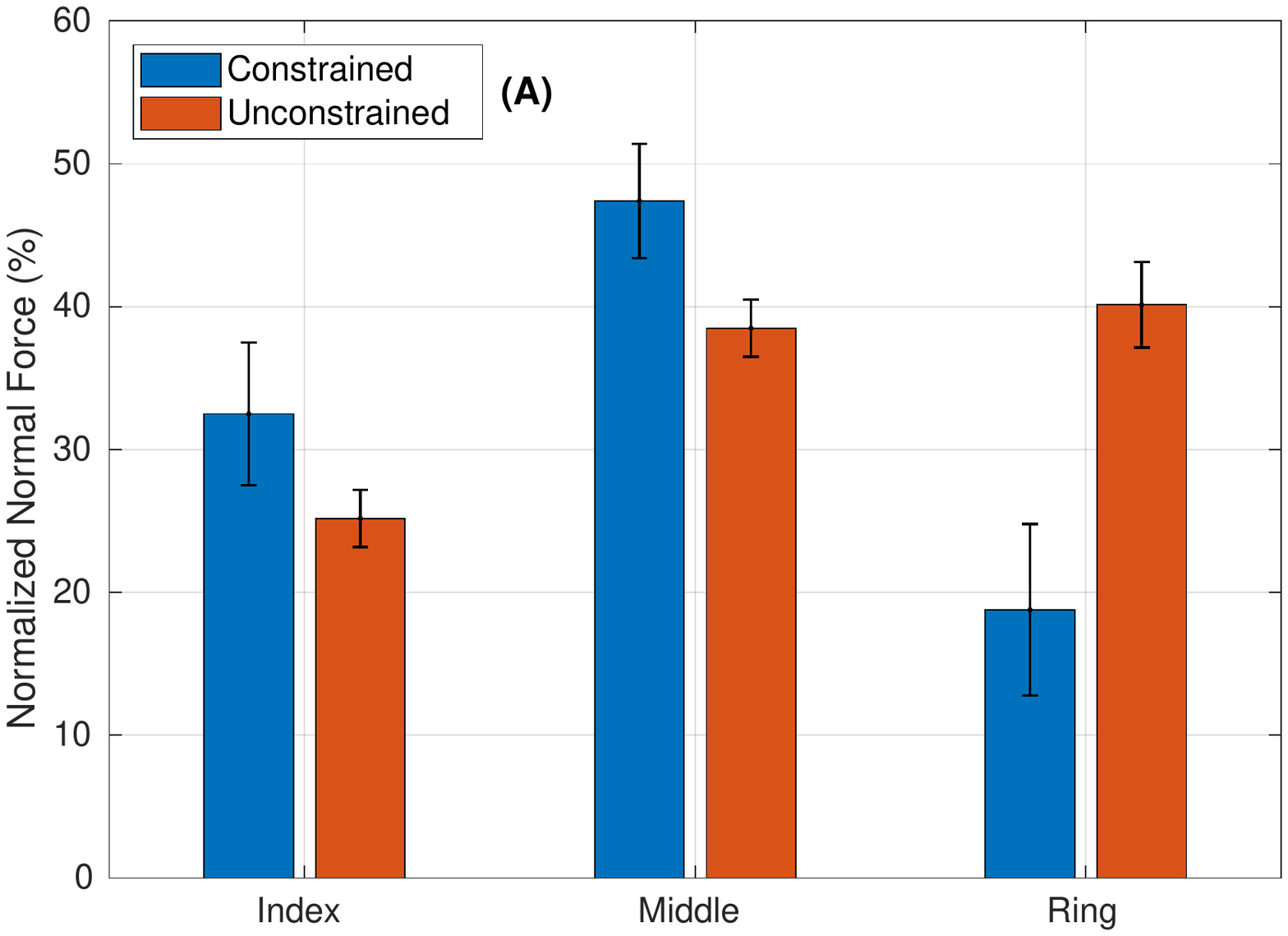}
  \vspace{-1em}
\end{figure}

\begin{figure}[!t]
      \centering
      \includegraphics[trim={2.5cm 8cm 2.4cm 7.9cm}, width=0.35\textwidth]{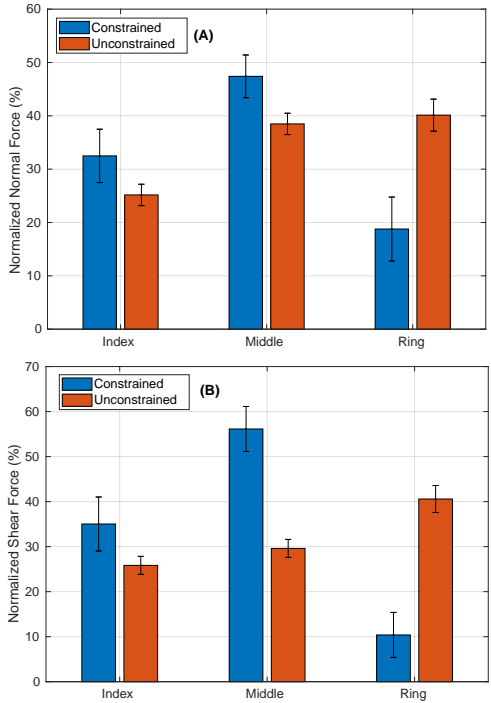}
      \caption{ The average normalized forces over the 10-second hold phase for the three fingers under the two grasping conditions. (a) The normalized estimated normal forces in percentage, (b) The normalized estimated shear forces in percentage.}
      \label{normalnormal}
      \vspace{-1.5em}
\end{figure}

The pairwise \textit{u} test has been performed on the 10-second average force across all 12 trials (2 trials per each subject) among all three fingers for the two different grasping conditions (Table~\ref{table}). The results indicate that there is a significant difference with $95\%$ confidence in the force magnitude when the grasping condition changes from constrained to unconstrained. However, the results indicate that there is no significant change in the total finger force or thumb force when the condition changes. The change in the thumb force is $0.35 \pm 0.02$ N for the normal and $0.27 \pm 0.01$ N for the shear forces, respectively (less than $3\%$ of the overall force range). This indicates that while the overall force required to grasp the object is not significantly affected by constraining the grasp, the distribution of forces among the individual fingers is. Moreover, It may be also seen from the error bars that there is more force variation in fingers across the subjects/trials for constrained grasping. Comparing the pairwise \textit{u} test result for the variance of each group, shows that there is a significant difference in each finger force variance between the two grasping conditions. In order to better explain the force disstribution change over the three fingers, each finger force has been normalized by the total force of each trial, and then the average normalized forces for three fingers are plotted as a percentage in Fig.~\ref{normalnormal}(a) and (b). It may be seen that the index finger ($7\%$ for normal and $9\%$ for shear) and the middle finger forces ($9\%$ for normal and $27\%$ for shear) have decreased, while the ring finger force ($21\%$ for normal and $30\%$ for shear) has increased by switching to the unconstrained condition. 

\begin{figure}[!t]
      \vspace{1em}

  \centering
  \includegraphics[trim={2cm 8.2cm 2cm 6cm}, width=0.35\textwidth]{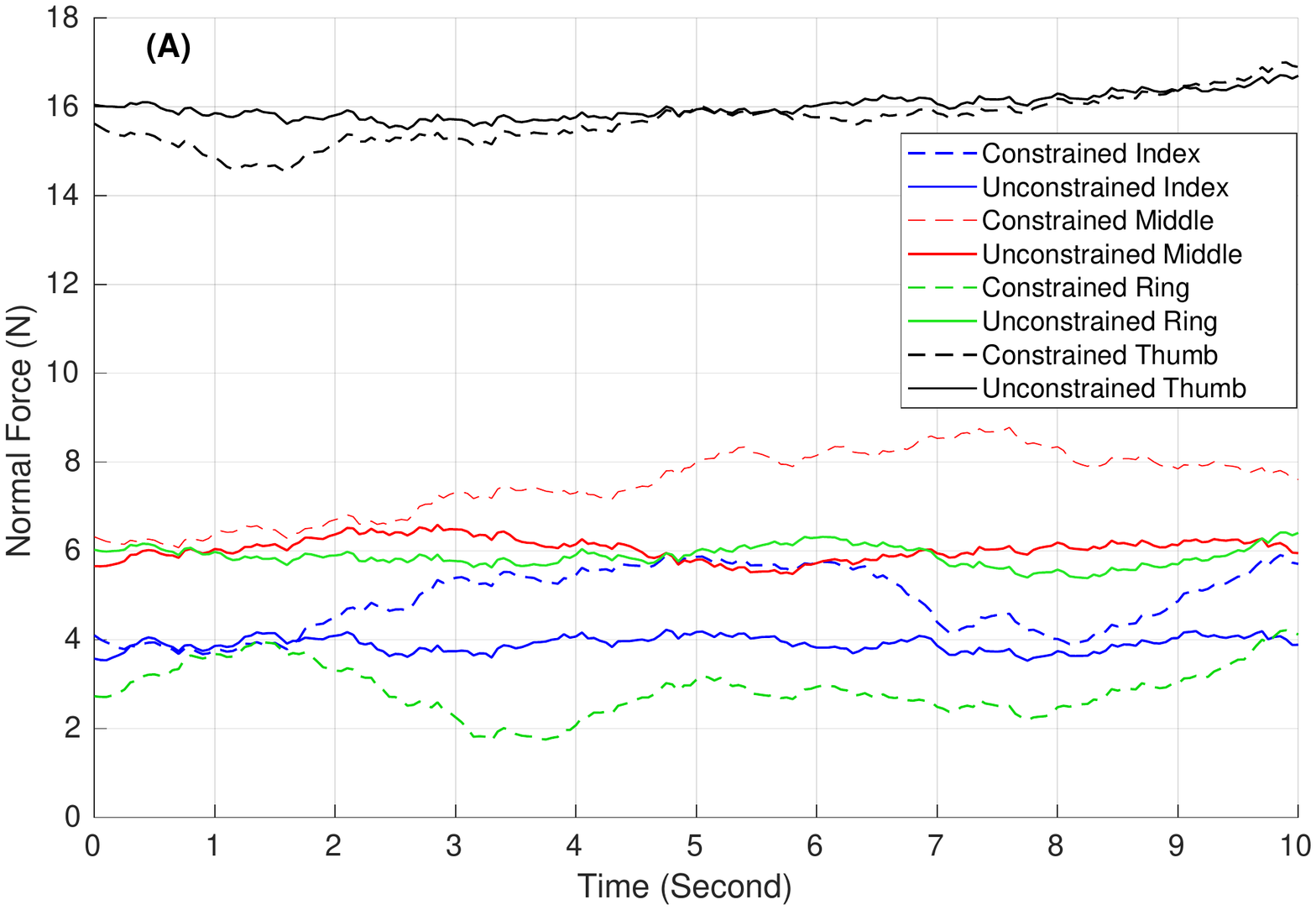}
  \vspace{-1em}
\end{figure}

\begin{figure}[!t]
      \centering
      \includegraphics[trim={2cm 7cm 2cm 6cm}, width=0.35\textwidth]{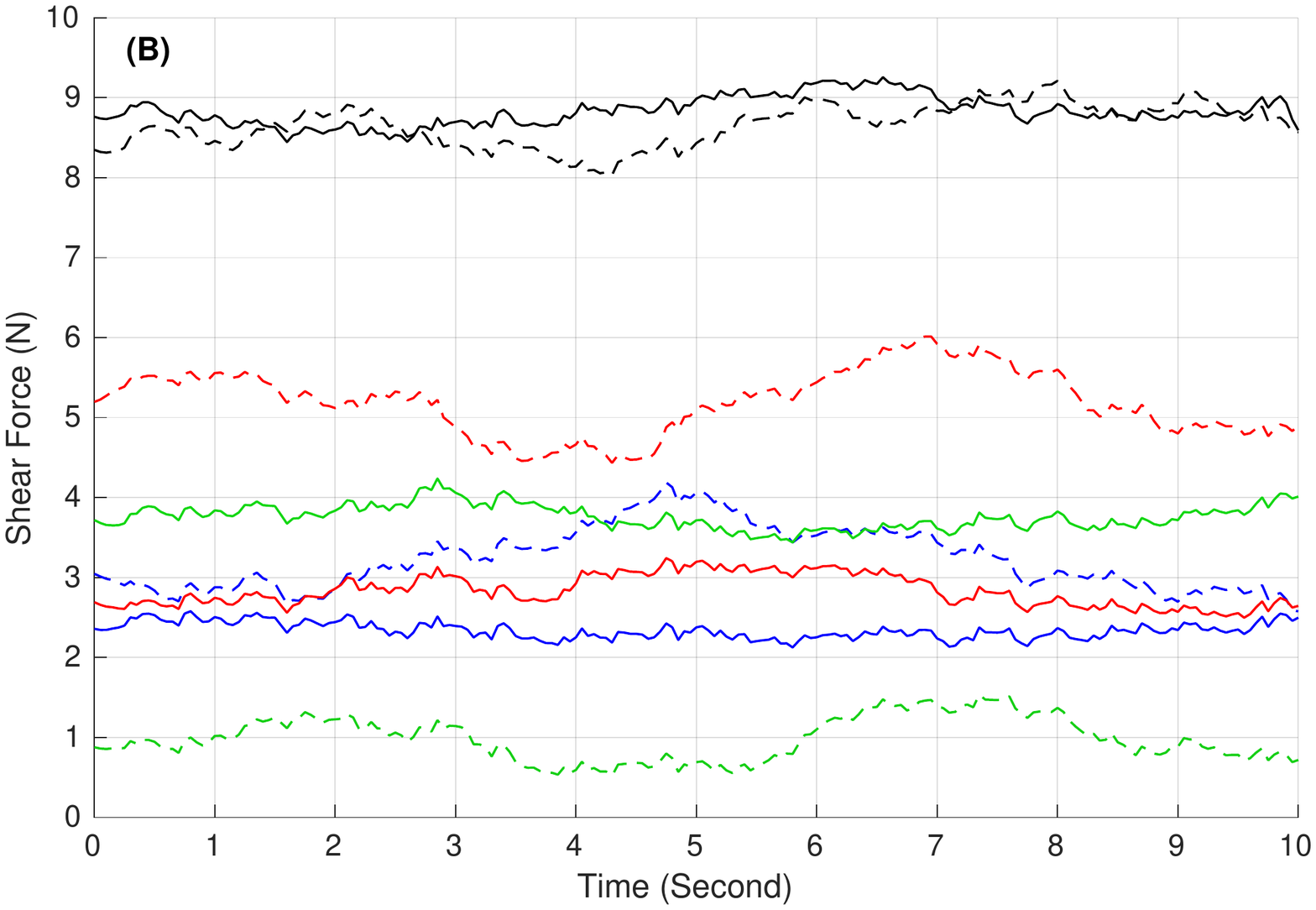}
      \caption{ The average estimated force per each finger during the hold phase under the two grasping conditions. (a) The average estimated normal forces, (b) The average estimated shear forces.}
      \label{normaltime}
      \vspace{-1.5em}
\end{figure}

\renewcommand{\arraystretch}{1.5}
\begin{table}[!b]
  \vspace{-1.5em}
  \centering
\caption{The pairwise \textit{u} test results performed on the average forces across all 12 trials for 2 grasping conditions.}
\label{table}
\begin{tabular}{ccc}
\toprule[1.5pt]

                  &     \textbf{Normal Force}     &\textbf{Shear Force}             \\ \midrule[1.5pt]
{\textbf{Index}}  & $p = 0.0022$, $\alpha = 0.05$ & $p = 0.0087$, $\alpha = 0.05$   \\  \hline
{\textbf{Middle}} & $p = 0.0046$, $\alpha = 0.05$ & $p = 0.0022$, $\alpha = 0.05$   \\  \hline
{\textbf{Ring}}   & $p = 0.0043$, $\alpha = 0.05$ & $p = 0.0031$, $\alpha = 0.05$   \\  \hline
{\textbf{Thumb}}  & $p = 0.5878$, $\alpha = 0.05$ & $p = 0.1320$, $\alpha = 0.05$   \\ 

\bottomrule[1.5pt]
\end{tabular}
\end{table}
\renewcommand{\arraystretch}{1}

 From Fig.~\ref{normalnormal}, it can be observed that the forces are more evenly distributed among the three fingers for unconstrained grasping. Force variance (across the fingers) in each trial is considered as a metric of balance. The average force variances  are found as $\bar{S}_{normal}^{C} = 4.62 \text{N}^2$ and $\bar{S}_{shear}^{C} = 5.29 \text{N}^2$ for constrained condition, and $\bar{S}_{normal}^{U} = 1.54 \text{N}^2$ and $\bar{S}_{shear}^{U} = 0.65 \text{N}^2$ for unconstrained condition. the pairwise \textit{u} test for the two groups of constrained and unconstrained variances shows that there is a significant difference with $95\%$ confidence in force balance among the three fingers.

\begin{figure}[!t]
  \centering
  \includegraphics[trim={2cm 8.1cm 2cm 6cm}, width=0.35\textwidth]{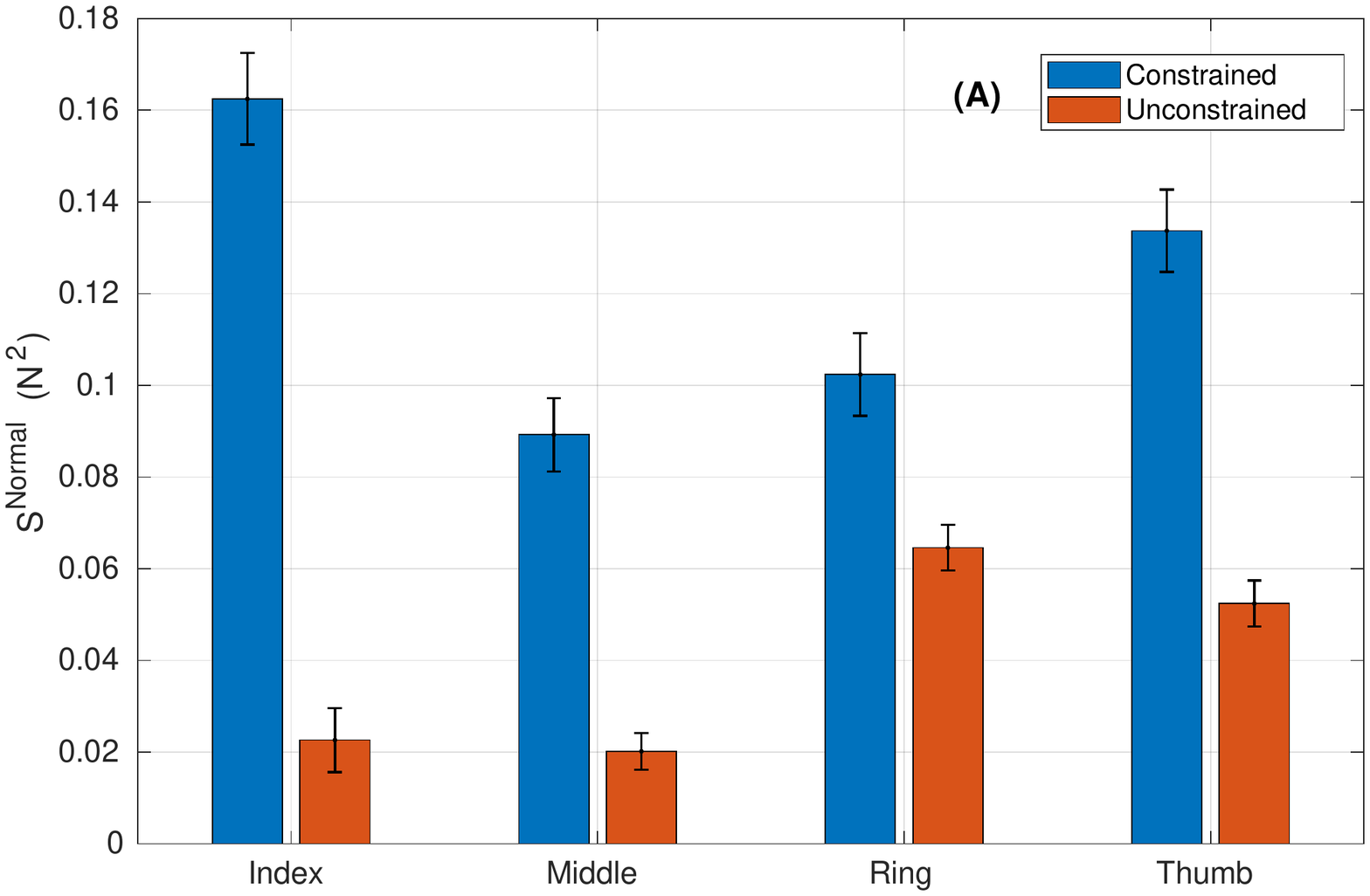}
  \vspace{-0.1em}
\end{figure}

\begin{figure}[!t]
      \centering
      \includegraphics[trim={2cm 7cm 2cm 7cm}, width=0.35\textwidth]{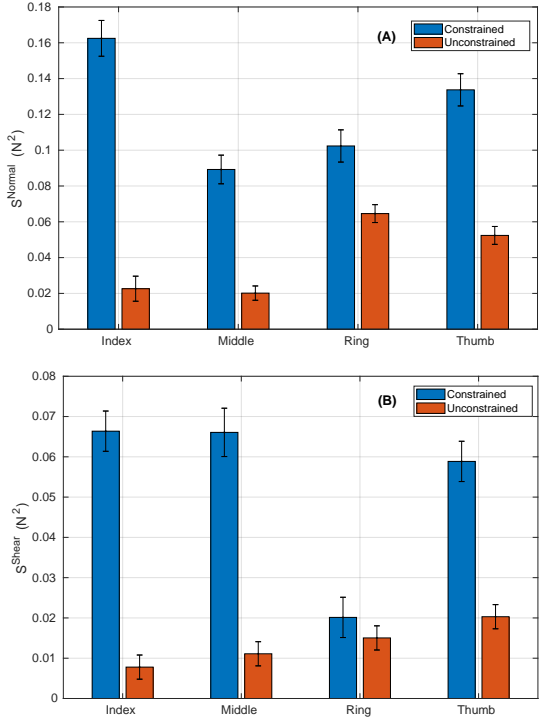}
      \caption{ The average force varinace per each finger across all human subjects. (a) The average normal forces varinace $\bar{S}_{i, i=1:4}^{normal}$, (b) The average normal forces varinace $\bar{S}_{i, i=1:4}^{shear}$.}
      \label{normalvar}
      \vspace{-1.5em}
\end{figure}

Another interesting result from the grasping tests is the steadiness of the applied forces during the hold phase in each finger. Fig.~\ref{normaltime}(a) and (b) show the average force variation over the 10 seconds period for each individual finger under the two grasping conditions. It can be observed that the forces are more steady during the unconstrained test. The variance (over time) of each finger force during the hold phase in each trial is considered as a metric of steadiness (${S}_{ij, i=1:4, j=1:12}^{normal}$ and ${S}_{ij, i=1:4, j=1:12}^{shear}$). Fig.~\ref{normalvar}(a) and (b) compare the average of these variances ($\bar{S}_{i, i=1:4}^{normal}$ and $\bar{S}_{i, i=1:4}^{shear}$) for the two grasping conditions. The result of the \textit{u} test for the two groups of constrained and unconstrained variances indicates that there is a significant difference with $95\%$ confidence in force steadiness by switching from constrained to the unconstrained condition.

\section{CONCLUSIONS}

This paper demonstrates the effectiveness of the fingernail imaging method in studying the unconstrained grasp forces as well as the effect of grasping constraints on force synergy among fingers. To overcome the problem of keeping fingers in the field of view of the cameras, two six degree-of-freedom robotic arms with the eye-in-hand camera setup and a visual servoing tracking system have been devised to ensure that there always exists a proper view of the human fingernails during the grasping experiments. The preliminary experimental results show the accuracy of force measurement based on fingernail imaging for estimating dynamic grasping forces with a maximum RMS error of $0.82 \pm 0.06$ N force over the four fingers, which is comparable to prior fingernail imaging work on quasi-static forces \cite{grieve20153,grieve2016optimizing}. 

The experimental results have also demonstrated a never-before-seen comparison of constrained vs. unconstrained grasp force distributions during a dynamic grasping task. Based on the experimental results, the main conclusions are: (1) Although the total grasp force is not significantly affected, the individual finger forces are significantly different for the two grasping conditions. The index and the middle finger forces have decreased, while the ring finger force has increased by switching to the unconstrained condition. (2) The forces are more evenly distributed (more balanced) among the three fingers under the unconstrained condition compared to the constrained grasping. (3) There is more consistency/repeatability in finger forces across the subjects/trials for unconstrained grasping. (4) By looking just at a single finger, it can be concluded that there is more force variation in constrained grasping over time. In other words, the unconstrained forces are more steady versus time. The combined observations that unconstrained grasp forces are more evenly distributed, repeatable, and steady is strong evidence that it is important to not constrain grasping in order to accurately study the natural grasping synergies of humans.

Future work will include the study of dynamic force behaviors during a manipulation task, which consists of both grasping and positioning stages. Additionally, the effect of changing the object center of mass on digit placement both before and during the grasping task will be studied in future research. Finally, it is the authors' goal to design a system of multiple collaborative robots to interactively work with a human subject in a workspace based on the estimated applied finger forces, while tracking their hands.

      \vspace{1em}

\addtolength{\textheight}{-12cm}  



\bibliographystyle{ieeetr}
\bibliography{ICRAref}

\end{document}